%% file: iclr2025_conference.tex
\documentclass{article} %
\usepackage{iclr2025_conference,times}
\input{math_commands.tex}

\usepackage{amsmath}
\usepackage{xcolor}
\usepackage{subcaption}
\usepackage{hyperref}
\usepackage{url}
\usepackage{booktabs}       %
\usepackage{graphicx}       %
\usepackage{amsfonts}       %
\usepackage{amsmath}
\usepackage{multirow}
\usepackage{nicefrac}       %
\usepackage{microtype}      %
\usepackage{romannum}
\usepackage{wrapfig}

\title{Cavia: Camera-controllable Multi-view Video Diffusion with View-Integrated Attention}

{
\author{
\textbf{
  \noindent Dejia Xu\textsuperscript{1}\thanks{This work was performed while Dejia Xu interned at Apple.}\,\,,Yifan Jiang\textsuperscript{2}, Chen Huang
\textsuperscript{2},
  Liangchen Song\textsuperscript{2}, Thorsten Gernoth\textsuperscript{2}}\\
\quad\quad\textbf{Liangliang Cao\textsuperscript{3}\textdagger, Zhangyang Wang\textsuperscript{1}, Hao Tang\textsuperscript{2}}\\
  {\textsuperscript{1}University of Texas at Austin, \textsuperscript{2}Apple, \textsuperscript{3}Google}
}
}

\iclrfinalcopy %
\begin{document}

\makeatletter
\newcommand{\conditionalFootnote}[1]{%
  \@ifundefined{iclrfinalcopy}{%
  }{%
\renewcommand{\thefootnote}{\fnsymbol{footnote}}
    \footnotetext[2]{#1}%
\renewcommand{\thefootnote}{\arabic{footnote}}
  }%
}
\makeatother
\maketitle
\conditionalFootnote{This work was performed while Liangliang Cao worked at Apple.}
\renewcommand{\thefootnote}{\arabic{footnote}}
\pagenumbering{arabic}

\begin{abstract}
In recent years there have been remarkable breakthroughs in image-to-video generation.
However, the 3D consistency and camera controllability of generated frames have remained unsolved. Recent studies have attempted to incorporate camera control into the generation process, but their results are often limited to simple trajectories or lack the ability to generate consistent videos from multiple distinct camera paths for the same scene. To address these limitations, we introduce \textbf{Cavia}, a novel framework for camera-controllable, multi-view video generation, capable of converting an input image into multiple spatiotemporally consistent videos. Our framework extends the spatial and temporal attention modules into view-integrated attention modules, improving both viewpoint and temporal consistency. This flexible design allows for joint training with diverse curated data sources, including scene-level static videos, object-level synthetic multi-view dynamic videos, and real-world monocular dynamic videos. To our best knowledge, Cavia is the first of its kind that allows the user to precisely specify camera motion while obtaining object motion. 
Extensive experiments demonstrate that Cavia surpasses state-of-the-art methods in terms of geometric consistency and perceptual quality. Project Page: \url{https://ir1d.github.io/Cavia/}
\end{abstract}

\section{Introduction}

The rapid development of diffusion models has enabled significant advancements in video generative models. Early efforts have explored various approaches, either training a video model from scratch or by fine-tuning pre-trained image generation models with additional temporal layers~\citep{svd,wang2023videofactory,ho2022b,singer2022make,ho2022a,nan2024openvid}. The training data of these video models typically consist of a curated mixture of image~\citep{schuhmann2022laion} and video datasets~\citep{bain2021frozen,wang2023internvid,wang2023videofactory,nan2024openvid}. 
While substantial progress has been made in improving model architectures and refining training data, relatively little research has been conducted on the 3D consistency and camera controllability of generated videos.

\looseness=-1
To tackle this issue, several recent works~\citep{wang2023motionctrl,he2024cameractrl,bahmani2024vd3d,xu2024camco,hou2024training} have attempted to introduce camera controllability in video generation, aiming to ensure that generated frames adhere to viewpoint instructions, thereby improving 3D consistency. 
These works either enhance viewpoint control through better conditioning signals~\citep{wang2023motionctrl,he2024cameractrl,bahmani2024vd3d} or by utilizing geometric priors, such as epipolar constraints~\citep{xu2024camco} or explicit 3D representations~\citep{hou2024training}.
However, despite these efforts, the generated videos often lack precise 3D consistency or are restricted to static scenes with little to no object motion. Moreover, it remains challenging for monocular video generators to produce multi-view consistent videos of the same scene from different camera trajectories.

Since independently sampling multiple sequences often results in significantly inconsistent scenes, generating multiple video sequences simultaneously is desirable. However, this remains extremely challenging due to the scarcity of multi-view video data in the wild, leading to multi-view generations limited to inconsistent near-static scenes or synthetic objects.
A concurrent work, CVD~\citep{cvd}, builds on multi-view static videos~\citep{real10k} and warping-augmented monocular  videos~\citep{bain2021frozen}, but it can only generate videos with limited baselines, yielding inconsistent results when object motion is present. Another concurrent work, Vivid-ZOO~\citep{li2024vivid}, leverages dynamic objects from Objaverse~\citep{deitke2023objaverse} dataset and renders multi-view videos to train a video generator. 
However, due to limited data sources, their results are primarily object-centric frames from fixed viewpoints, lacking realistic backgrounds.

To address these challenges, we propose \textbf{Cavia}, a novel framework that extends a monocular video generator~\citep{svd} to generate multi-view consistent videos with precise camera control. We enhance the spatial and temporal attention modules to cross-view and cross-frame 3D attentions respectively, improving consistency across both viewpoints and frames.
Our model architecture enables a novel joint training strategy that fully utilizes static, monocular, and multi-view dynamic videos. Static videos~\citep{real10k,yu2023mvimgnet,xia2024rgbd,reizenstein2021common,deitke2023objaverse,deitke2023objaverse2} are converted to multi-view formats to ensure the geometric consistency in the generated frames. We then incorporate rendered synthetic multi-view videos of dynamic 3D objects~\citep{liang2024diffusion4d,jiang2024animate3d,li2024puppet} to teach the model to generate reasonable object motion. To prevent overfitting on synthetic data, we finetune the model on pose-annotated monocular videos~\citep{wang2023internvid,nan2024openvid} to enhance performance on complex scenes.
Our framework synthesizes cross-view and cross-frame consistent videos, and extensive evaluations on real and text-to-image generated images show its applicability across challenging indoor, outdoor, object-centric, and large-scene cases. We systematically measure the quality of the generated videos in terms of per-video and cross-view geometric consistency and perceptual quality. Our experiments demonstrate our superiority compared to previous works both qualitatively and quantitatively. Our experiments demonstrate superior performance compared to previous methods, both qualitatively and quantitatively. Additionally, we show that our method can extrapolate to generate four views during inference and enable 3D reconstruction of the generated frames.

Our main contributions can be summarized as follows,

\begin{itemize}
    \item We propose a novel framework, \textbf{Cavia}, for generating multi-view videos with camera controllability. We introduce view-integrated attentions, namely cross-view and cross-frame 3D attentions, to enhance consistency across viewpoints and frames.
    \item We introduce an effective joint training strategy that leverages a curated mixture of static, monocular dynamic, and multi-view dynamic videos, ensuring geometric consistency, high-quality object motion, and background preservation in the generated results.
    \item Our experiments demonstrate superior geometric and perceptual quality in both monocular video generation and cross-video consistency compared to baseline methods. Additionally, our flexible framework can operate on four views at inference, offering improved view consistency and enabling 3D reconstruction of the generated frames.
\end{itemize}

\section{Related Works}

\subsection{Camera Controllable Video Diffusion Models}
\looseness=-1
Recent advancements in video diffusion models have significantly benefited from scaling model architectures and leveraging extensive datasets~\citep{bain2021frozen,wang2023internvid,wang2023videofactory}, leading to impressive capabilities in generating high-quality videos~\citep{svd,ho2022b,singer2022make,ho2022a,sora2023}.
While large foundational video diffusion models exist, our work focuses on enhancing camera control over video diffusion processes, a rapidly growing area of research. AnimateDiff~\citep{guo2023animatediff} and Stable Video Diffusion (SVD)~\citep{svd} employ individual camera LoRA~\citep{hu2021lora} models for specific camera motions. MotionCtrl~\citep{wang2023motionctrl} improves flexibility by introducing camera matrices, while CameraCtrl~\citep{he2024cameractrl}, CamCo~\citep{xu2024camco}, and VD3D~\citep{bahmani2024vd3d} enhance the camera control accuracy by introducing Plücker coordinates to the video models via controlnet~\citep{controlnet}. To further improve the geometric consistency, CamCo~\citep{xu2024camco} applies epipolar constraints and CamTrol~\citep{hou2024training} incorporates 3D Gaussians~\citep{kerbl20233d}. However, these methods focus on monocular video generation, limiting their ability to sample multiple consistent video sequences of the same scene from distinct camera paths. CVD~\citep{cvd} extends CameraCtrl~\citep{he2024cameractrl} for multi-view video generation, but their results are constrained to simple camera and object motion. ViVid-Zoo~\citep{li2024vivid} extends MVDream~\citep{shi2023mvdream} for multi-view purposes but is limited to object-centric results with fixed viewpoints. In contrast, our work explores view-integrated attentions for more precise camera control over arbitrary viewpoints and introduces a joint training strategy leveraging data mixtures to improve novel-view performance in complex scenes.

\subsection{Multi-view Image Generation}
Early approaches such as MVDiffusion \citep{Tang2023mvdiffusion} focused on generating multiview images in parallel by employing correspondence-aware attention mechanisms, enabling effective cross-view information interaction, particularly for textured scene meshes. Recent approaches like Zero123++ \citep{shi2023zero123++}, Direct2.5 \citep{lu2024direct2}, Instant3D \citep{instant3d2023}, MVDream \citep{shi2023mvdream}, MVDiffusion++ \citep{tang2024mvdiffusion++}, CAT3D \citep{gao2024cat3d}, and Wonder3D \citep{long2024wonder3d} have introduced single-pass frameworks for multiview generation, utilizing multiview self-attention to improve viewpoint consistency. Other works, such as SyncDreamer \citep{liu2023syncdreamer}, One-2-3-45 \citep{liu2024one}, Cascade-Zero123 \citep{chen2023cascade} and ConsistNet \citep{yang2024consistnet}, incorporate multiview features into 3D volumes to facilitate 3D-aware diffusion models \citep{liu2023zero,watson2022novel}. 
Meanwhile, techniques such as Pose-Guided Diffusion \citep{poseguideddiffusion}, Era3D \citep{li2024era3d}, Epidiff \citep{huang2024epidiff}, and SPAD \citep{kant2024spad} have integrated epipolar-based features to facilitate enhanced viewpoint fusion within diffusion models. Finally, approaches like V3D \citep{chen2024v3d}, IM-3D \citep{melaskyriazi2024im3d}, SV3D \citep{voleti2024sv3d} and Vivid-1-to-3 \citep{kwak2024vivid} leverage priors from video diffusion models to achieve multiview generation with improved consistency. 
However, these methods focus on generating static 3D objects or scenes, while our work introduces vivid object motion into multiview dynamic video generation in complex scenes.

\subsection{4D Generation}
Recent efforts in 4D generation have explored various methods~\citep{mav3d, zhao2023animate124,4dfy,zheng2023unified,ling2023align} that use score distillation from video diffusion models to optimize dynamic NeRFs or 3D Gaussians for text- or image-conditioned scenes.  
Follow-up works~\citep{jiang2023consistent4d,ren2023dreamgaussian4d,yin20234dgen,ren2024l4gm,zeng2024stag4d,pan2024fast} investigate video-to-4D generation, enabling controllable 4D scene generation from monocular videos.
More recent methods~\citep{liang2024diffusion4d,xie2024sv4d,zhang20244diffusion} utilize video diffusion models to address the spatial-temporal consistency required for efficient 4D generation. 
However, these approaches primarily focus on object-centric generation and face challenges in producing realistic results with complex backgrounds. In contrast, our work emphasizes generating multi-view, 3D-consistent videos for complex scenes.

\section{Method}

\subsection{Overview}

\begin{figure}
    \centering
    \includegraphics[width=0.9\linewidth]{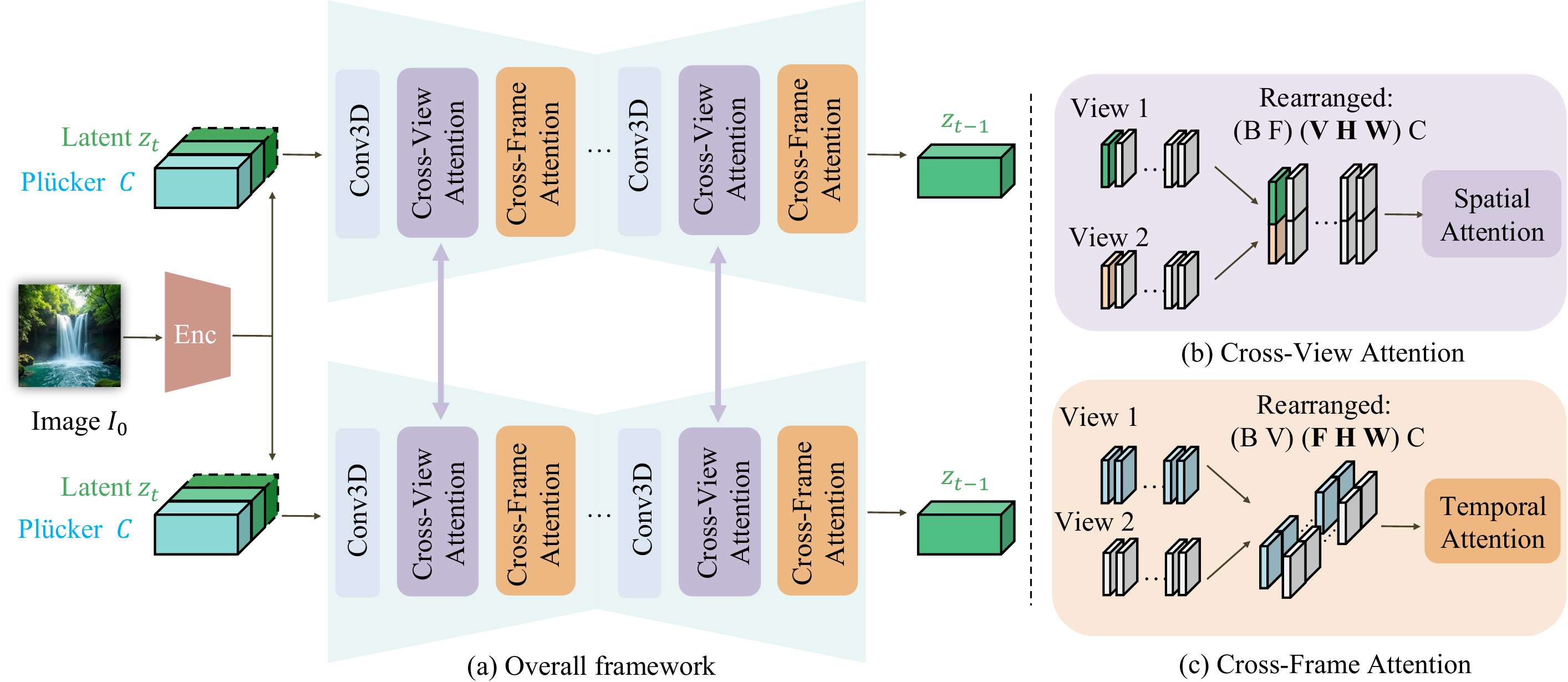}
    \caption{An overview of Cavia is shown in (a). We introduce view-integrated attention modules, namely cross-view attentions and cross-frame attentions, which enforce viewpoint and temporal consistency of the generated frames, respectively. As illustrated in (b) and (c), our view-integrated attention incorporates additional feature dimensions into the attention mechanism, enhancing consistency across views and frames.}
    \label{fig:arch}
\end{figure}

Image-to-video generation takes a single image $I_0$ as input and outputs a video sequence $O_1, \cdots, O_n$. By introducing camera control, the model additionally takes in a sequence of camera information $C_1, \cdots, C_n$, which dictates the desired viewpoint changes for the output sequence. In the multi-view scenario, we extend each batch of the camera control signal and output video sequence to $V$ sequences.
In the following paragraphs, we present our proposed \textbf{Cavia} framework in detail. 
First, we outline the preliminaries of image-to-video diffusion and describe how camera controllability is introduced in monocular video generation. Then, we elaborate on the model design for multi-view consistent video generation.
An overview of our framework is provided in Fig.~\ref{fig:arch}.

\subsection{Camera Controllable Video Diffusion Model}

\paragraph{Preliminaries}

Our model builds on pre-trained Stable Video Diffusion (SVD)~\citep{svd}. 
SVD extends Stable Diffusion 2.1\citep{rombach2022high} by adding temporal convolution and attention layers, following the VideoLDM architecture~\citep{videoldm}.
SVD is trained with a continuous-time noise scheduler~\citep{karras2022elucidating}. In each iteration, the training data is perturbed by Gaussian noise $\mathbf{n}(t) \sim \mathcal{N}(0, \sigma^2(t)\mathbf{I})$ and the diffusion model is tasked with estimating the clean data $x_0 \sim p_0$. Let $p(\mathbf{x} ; \sigma(t))$ denote the marginal probability of noisy data $\mathbf{x}_t = \mathbf{x}_0 + \mathbf{n}(t)$, the iterative refinement process of diffusion model corresponds to the probability flow ordinary differential equation (ODE):
\begin{equation}
d \mathbf{x}=-\dot{\sigma}(t) \sigma(t) \nabla_{\mathbf{x}} \log p(\mathbf{x} ; \sigma(t)) dt.
\end{equation}
$ \nabla_{\mathbf{x}} \log p(\mathbf{x}; \sigma(t))$ refers to the score function, which is parameterized by a denoiser $D_{\boldsymbol{\theta}}$ through
$\nabla_{\mathbf{x}} \log p(\mathbf{x} ; \sigma) \approx \left(D_{\boldsymbol{\theta}}(\mathbf{x} ; \sigma)-\mathbf{x}\right) / \sigma^2$. We follow the EDM-preconditioning framework~\citep{karras2022elucidating,svd} and parameterize $D_{\boldsymbol{\theta}}$ with a neural network $F_{\boldsymbol{\theta}}$ as follows,
\begin{equation}
    D_{\boldsymbol{\theta}} = c_\text{skip} \mathbf{x} + c_\text{out} F_{\boldsymbol{\theta}} (c_\text{in} \mathbf{x}; c_\text{noise}).
\end{equation}
During training, the network $F_{\boldsymbol{\theta}}$ is optimized using denoising score matching for $D_{\boldsymbol{\theta}}$:
\begin{equation}
\mathbb{E}\left[\left\|D_{\boldsymbol{\theta}}\left(\mathbf{x}_0+\mathbf{n} ; \sigma, \text{cond}\right)-\mathbf{x}_0\right\|_2^2\right].
\end{equation}

\paragraph{Camera Conditioning}
Although SVD is pre-trained on various high-quality video and image data, it does not natively support precise camera control instructions directly.
To address this, we introduce camera conditioning to the model via Plücker coordinates ~\citep{jia2020plucker}, which is widely adopted as position 
 embeddings in 360$^\circ$ unbounded light fields\citep{sitzmann2021light}. Plücker coordinates are defined as $P = (d', o \times d')$, where $\times$ is the cross product and $d'$ refers to the normalized ray direction $d'=\frac{d}{||d||}$. Let camera extrinsic matrix be $E=[\bf{R}|\bf{T}]$ and intrinsic matrix be $\bf{K}$, 
the ray direction $d_{x, y}$ for 2D pixel located at $(x, y)$ is formulated as $d=\bf{R}\bf{K}^{-1}(\begin{smallmatrix}x\\ y\\ 1\end{smallmatrix})+\bf{T}$.
These spatial Plücker coordinates are concatenated channel-wise with the original latent inputs of SVD. We enlarge the convolution kernel of the first layer accordingly. The newly introduced matrices are zero-initialized to ensure training stability.

We utilize a relative camera coordinate system, where the first frame is positioned at the world origin with an identity matrix for rotation. The following frames are rotated accordingly. To stabilize training, we normalize the scale of the training sequences to a unit scale. This is implemented by resizing the maximum distance-to-origin in the multi-view camera sequence to 1.

\paragraph{Cross-frame Attention for Temporal Consistency}

Vanilla 1D temporal attention in the SVD backbone is insufficient for modeling large pixel displacements when the viewpoint changes~\citep{shi2023mvdream,yang2024cogvideox}. In vanilla 1D temporal attention, attention matrices are calculated over the frame number dimension, and latent features only interact with features from the same spatial location across frames. This limits information flow between different spatial-temporal locations. While this might not be a big issue for video generation with limited motion, viewpoint changes typically cause significant pixel displacements, which calls for better architecture for more efficient information propagation.

To overcome this issue, we inflate the original 1D temporal attention modules in the SVD network into 3D cross-frame temporal attention modules, allowing for joint modeling of spatial-temporal feature coherence. 
The inflation operation can be achieved by rearranging the latent features before the attention matrix calculations. 
Consider the latent features of shape $(B\hspace{0.25em} V\hspace{0.25em} F\hspace{0.25em} C\hspace{0.25em} H\hspace{0.25em} W)$ where $F$ refers to the length of frames and $V$ is the number of views, instead of employing 1D attention mechanism on rearranged shape of $((B\hspace{0.25em} V\hspace{0.25em} H\hspace{0.25em} W)\hspace{0.25em} F\hspace{0.25em} C)$, our inflated attention operates on the rearranged shape of $((B\hspace{0.25em} V)\hspace{0.25em} (F\hspace{0.25em} H\hspace{0.25em} W)\hspace{0.25em} C)$, integrating spatial features into the attention matrices.  
A visualization is provided in Fig.~\ref{fig:arch}(c).

Since our rearrange operation only alters the sequence length of the attention inputs without modifying the feature dimensions, we can seamlessly inherit the pre-trained weights from the SVD backbone for our purpose. Thanks to this rearrange operation, our inflated temporal attention now calculates the similarity of spatial-temporal features simultaneously, accommodating larger pixel displacements while maintaining temporal consistency.

\subsection{Consistent Multi-view Video Diffusion Model}

Adding Plücker coordinates for camera control and introducing improved temporal attention allows the video diffusion model to generate reasonably consistent monocular videos. However, for multi-view generation, a monocular video diffusion model that generates samples independently cannot ensure view consistency across multiple sequences. To address this, we introduce novel design mechanisms and training strategies to extend the monocular video diffusion model to the multi-view generation task.

\paragraph{Cross-view Attention for Multi-view consistency}

To improve cross-view consistency in multi-view videos, we aim to encourage information exchange during the generation process. 
Since our temporal cross-frame attention modules already handle intra-view feature connections within each video sequence, we focus on exchanging inter-view signals through the spatial cross-view modules.
Inspired by MVDream~\citep{shi2023mvdream}, we introduce 3D cross-view attention modules, inflated from the spatial attention blocks of SVD~\citep{svd}. 
Specifically, we rearrange the \texttt{V} views such that frames at each corresponding timesteps are concatenated before being sent into the attention modules.
In detail, we rearrange the latent features from shape $(B\hspace{0.25em} V\hspace{0.25em} F\hspace{0.25em} C\hspace{0.25em} H\hspace{0.25em} W)$ to $(((B\hspace{0.25em} F) (V\hspace{0.25em} H\hspace{0.25em} W)\hspace{0.25em} C)$ instead of $(((B\hspace{0.25em} V\hspace{0.25em} F) (H\hspace{0.25em} W)\hspace{0.25em} C)$. A visualization is provided in Fig.~\ref{fig:arch}(b).

Since only the second-to-last dimension, representing token length, is extended while other dimensions remain unchanged, our inflated spatial attention can inherit the model weights from the monocular setting. This flexibility allows our model to leverage training data with varying numbers of views and facilitates extrapolation to additional views at inference.
To handle multi-view generation, we introduce an additional view dimension to the input data.
To maintain workflow simplicity, we absorb the view dimension into the batch dimension during processing of other blocks, ensuring flexibility in handling different numbers of views.

\section{Joint Training Strategy on Curated Data Mixtures}

Thanks to the view-integrated attention mechanism, which allows for inheriting the module weights, our framework can leverage various data sources, including static, multi-view dynamic, and monocular videos. This is hard to achieve in previous methods. In this section, we first illustrate our joint training strategy, followed by details on the curated data mixtures.

\subsection{Joint Training on Data Mixtures}
For videos capturing static scenes~\citep{real10k,yu2023mvimgnet,xia2024rgbd,reizenstein2021common,deitke2023objaverse,deitke2023objaverse2}, we consider all frames to be temporally synchronized. 
An arbitrary subsequence of length $(F-1) \times V + 1$ from the original video can be reformatted into a $V$-view sequence with a shared starting frame and $F$ total frames per view. Static scenes also allow frame order reversal, providing additional augmentation opportunities.
We further prepare multi-view dynamic videos by rendering animatable objects from Objaverse~\citep{liang2024diffusion4d,jiang2024animate3d}. We design random smooth trajectories with diverse elevation and azimuth changes to avoid overfitting on simple camera movements. These trajectories start from a shared random forward-facing starting point and result in $n \times v$ frames in total. 

To avoid the model overfitting on synthetic images with simple backgrounds, we include a portion of data from monocular in-the-wild videos~\citep{wang2023internvid,nan2024openvid}. Training multi-view camera control from monocular videos is extremely challenging. Although CamCo~\citep{xu2024camco} and 4DiM~\citep{watson2024controlling} have explored joint training for monocular video generation, these approaches are unsuitable for multi-view scenarios.  The concurrent work CVD~\citep{cvd} explored homography warping to augment the monocular videos into pseudo-multi-view videos, but the limited realism of these augmentations restricts their ability to generate complex camera and object motion.

To overcome these issues, we choose to jointly train our model on monocular and multi-view videos to effectively utilize the abundant object motion information from all data sources. We annotate the monocular videos with camera poses using Particle-SfM~\citep{zhao2022particlesfm}.
Since in-the-wild monocular videos often contain noisy or unnatural content, we apply a rigorous filtering pipeline to remove unsuitable clips.
These curated video clips, sourced from InternVid~\citep{wang2023internvid} and OpenVid~\citep{nan2024openvid} datasets, provide rich object motion as well as complex backgrounds that mitigate the gap between scene-level static data and object-level dynamic data.
We rearrange monocular videos as $V=1$ samples so that all data items can be processed uniformly without bells and whistles. Thanks to our view-integrated attention modules, which accommodate varying token lengths, the varying view numbers $V$ do not affect the training process.

\begin{figure}[t]
    \centering
    \includegraphics[width=\textwidth]{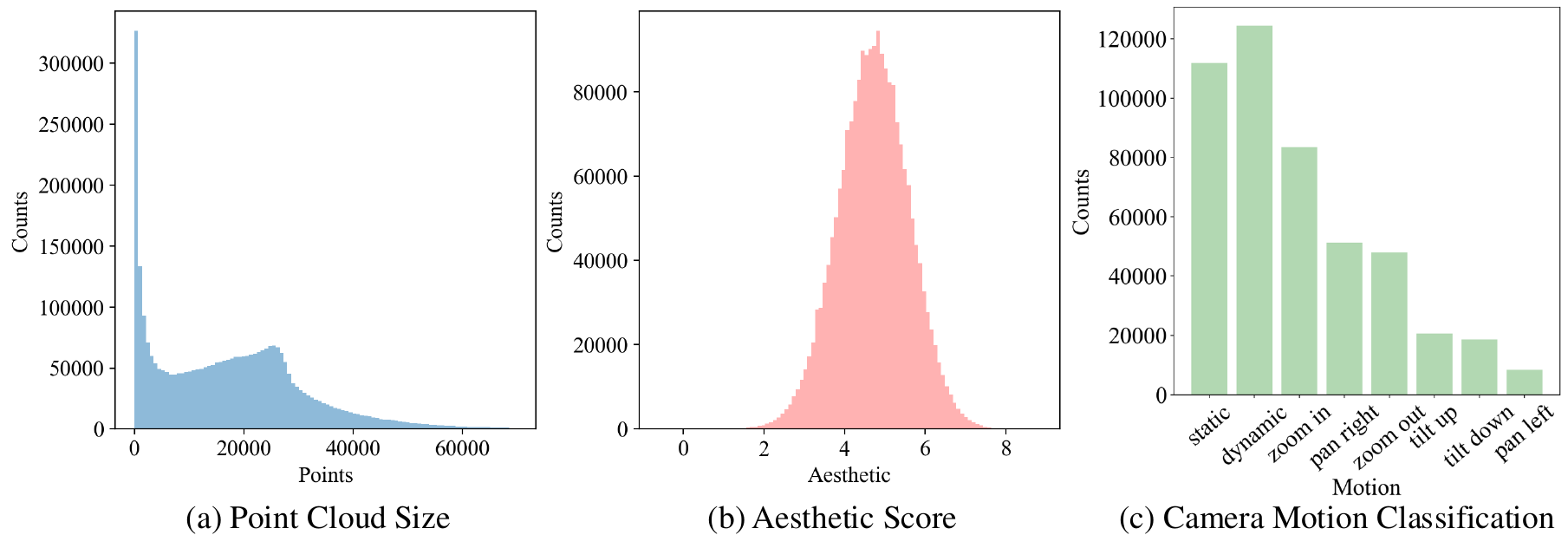}
    \caption{Statistics of the (a) point cloud size, (b) aesthetic score, and (c) camera motion classification result for our monocular video dataset.}
    \label{fig:filter}
    \vspace{-5mm}
\end{figure}

\begin{wrapfigure}{r}{0.5\textwidth} %
    \centering
        \includegraphics[width=0.4\textwidth]{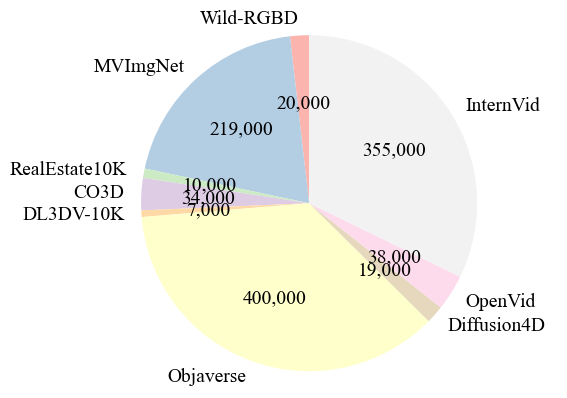}
    \caption{Sources of our training videos.}
    \vspace{-2mm}
    \label{fig:data_summary}
\end{wrapfigure}

\vspace{-3mm}
\subsection{Data Curation}
\vspace{-3mm}
\label{sec:data}
We begin by training our model extensively on static video data sourced from various publicly available datasets. Wild-RGBD~\citep{xia2024rgbd} includes nearly 20,000 RGB-D videos across 46 common object categories.
MVImgNet~\citep{yu2023mvimgnet} comprises 219,188 videos featuring objects from 238 classes. DL3DV-10K~\citep{ling2023dl3dv} provides 7,000 long-duration videos captured in both indoor and outdoor environments. CO3Dv2~\citep{reizenstein2021common} contains 34,000 turntable-like videos of rigid objects, crowd-sourced by nonexperts using cellphone cameras. Objaverse~\citep{deitke2023objaverse} and Objaverse-XL~\citep{deitke2023objaverse2} exhaustively crawl 10 million publicly available 3D assets. From these, we filtered out low-quality assets, such as those with incorrect textures or overly simplistic geometry, yielding a high-quality subset of 400,000 assets.

Similar to Diffusion4D~\citep{liang2024diffusion4d} and Animate3D~\citep{jiang2024animate3d}, we filter the animatable objects from Objaverse's Sketchfab subset. We exclude objects with excessive motion, which might result in partial observations, as well as nearly static objects with minimal motion. This curation process helps us obtain 19,000 high-quality dynamic assets that can be rendered from arbitrary viewpoints and timesteps, facilitating multi-view video generation. During each training iteration, we augment the frames with randomly selected background colors.

\looseness=-1
To improve the model's ability to generate object motion in the presence of complex backgrounds, we prepare monocular videos with camera pose annotations similar to CamCo~\citep{xu2024camco}. First, we use Particle-SfM~\citep{zhao2022particlesfm} to estimate the camera poses for randomly sampled frames from videos from InternVid~\citep{wang2023internvid} and OpenVid~\citep{nan2024openvid}. Inspired by CO3D~\citep{reizenstein2021common} and CamCo~\citep{xu2024camco}, we remove the videos where SfM fails to register all available frames or produces a point cloud with too few points or too many points.  
Fig.~\ref{fig:filter}(a) shows the point count statistics. A point cloud with too few points indicates poor frame registration to a shared 3D representation, while too many points suggest a mostly static scene, which is undesirable as we focus on object motion. Additionally, non-registered frames may indicate potential scene changes.
We then apply a rigorous filtering pipeline to ensure the quality of the video samples used for training.
This includes filtering based on aesthetic scores, optical character recognition (OCR), and camera motion classification using optical flow. Videos containing detected character regions are aggressively removed. Fig.~\ref{fig:filter}(b) and (c) present statistics on aesthetic score and camera motion classification results. Videos with low aesthetic scores or those classified as having static camera motion are excluded from the training set. Ultimately, we construct a dataset of 393,000 monocular videos annotated with camera poses. We provide a summary of the data sources used in Fig.~\ref{fig:data_summary}. More details and analysis are provided in the appendix.

\begin{figure}[h]
    \centering
    \vspace{-5mm}
    \includegraphics[width=0.95\linewidth]{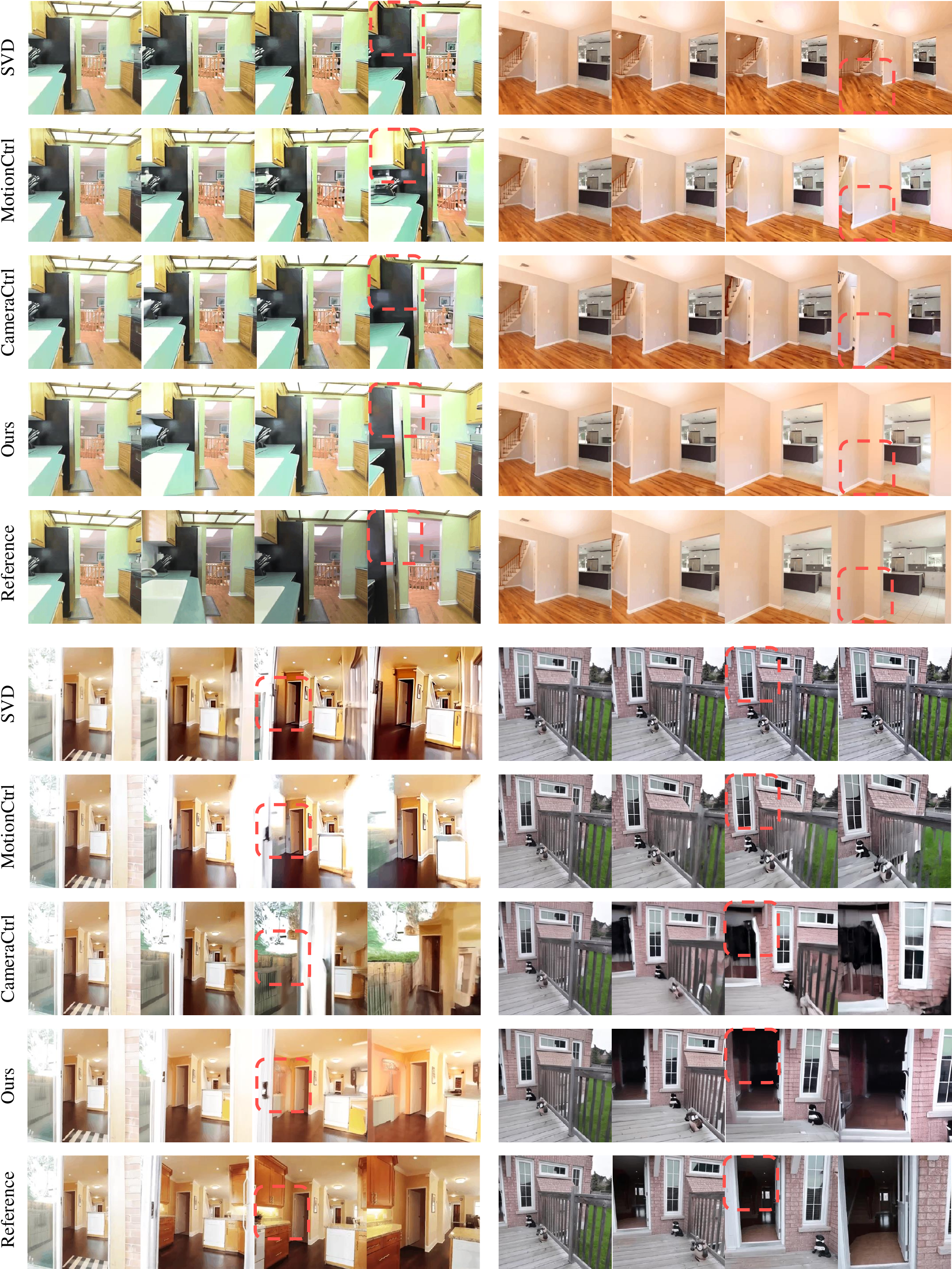}
    \caption{Per-video qualitative comparisons. The first frame in each reference set is the input image. Neither the image nor the camera trajectories were seen during model training. Video results are provided in supplementary for clearer qualitative comparisons.}
    \vspace{-5mm}
    \label{fig:mono}
\end{figure}

\section{Experiments}

In this section, we present experimental results and analysis. Video comparisons are included in the supplementary material for optimal visual evaluation. 
It is important to note that for all qualitative and quantitative evaluations, neither the input images nor the camera trajectories were used during model training.

\subsection{Quantitative Comparisons}

\paragraph{3D Consistency of Frames}
We evaluate the 3D consistency of the generated videos using COLMAP~\citep{schoenberger2016sfm,schoenberger2016mvs}. COLMAP is widely adopted for 3D reconstruction methods where camera pose estimation is required for in-the-wild images. We configure the COLMAP following previous methods~\citep{dsnerf, xu2024camco} for best few-view performance. 
A higher COLMAP error rate indicates poorer 3D consistency in the input images. Motivated by this, we report COLMAP errors as a measure of the 3D consistency of the frames. Each video is retried up to five times to reduce randomness.
We randomly sample 1,000 video sequences from RealEstate10K~\citep{real10k} test set for evaluation. 
Since we have ground truth 3D scenes, we use the ground truth camera pose sequences as the viewpoint instruction of the video model and compare the generated frames against the ground truth images.
Similar to prior works~\citep{he2024cameractrl, xu2024camco}, we extract the estimated camera poses and calculate the relative translation and rotation differences. 
Specifically, given two camera pose sequences, we convert them to relative poses and align the first frames to world origin. 
We then measure the angular errors in translation and rotation. Unlike previous works~\citep{he2024cameractrl,xu2024camco} that calculate the Euclidean distance of translation vectors, we use angular error measurements to ensure the camera pose scales are normalized, addressing scale ambiguity. As shown in Tab.~\ref{tab:comparison}, we calculate the area under the cumulative error curve (AUC) of frames whose rotation and translations are below certain thresholds ($5^\circ$, $10^\circ$, $20^\circ$). Our method significantly outperforms existing baselines.

\begin{figure}[h]
    \centering
    \includegraphics[width=\linewidth]{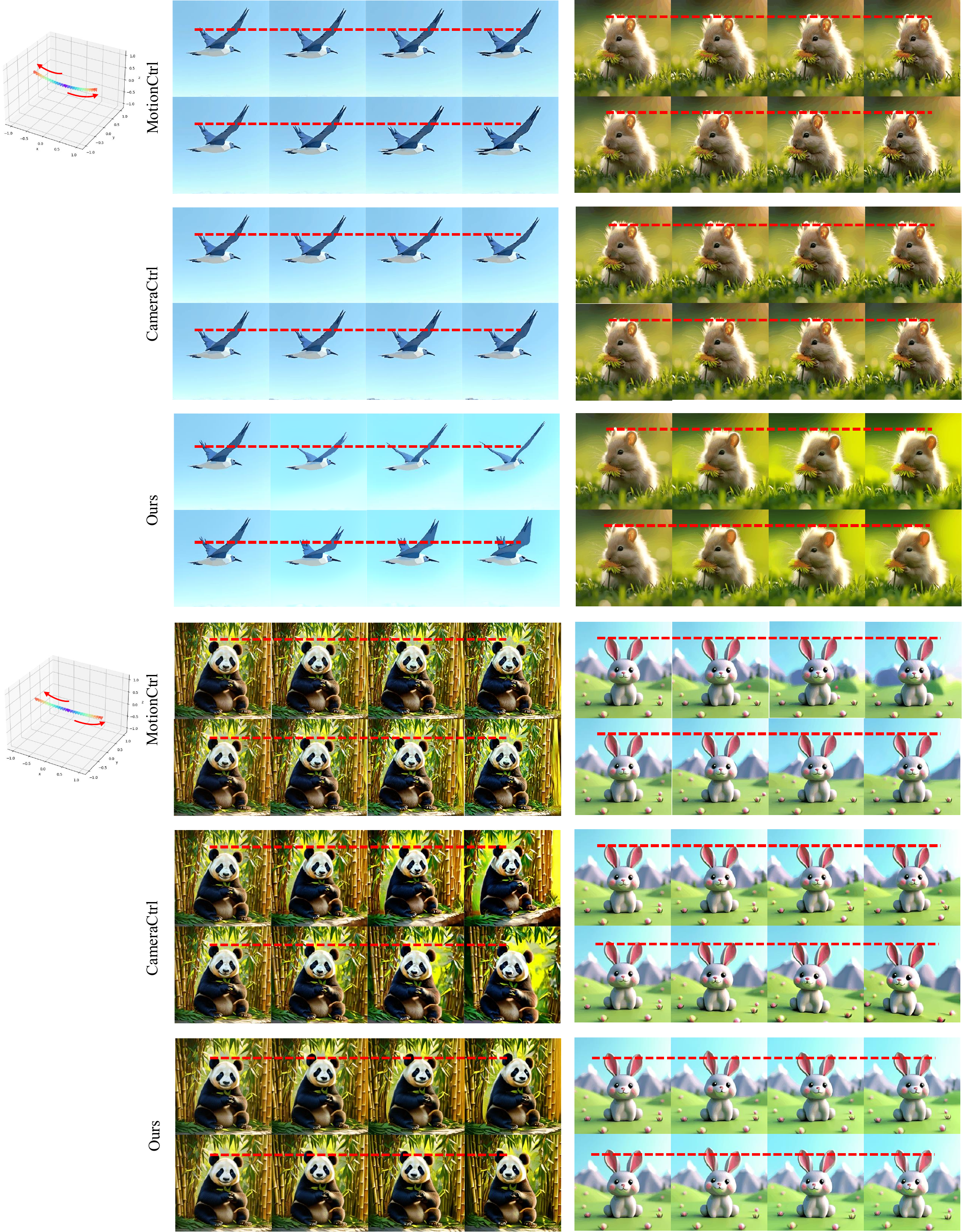}
    \vspace{-1mm}
    \caption{Qualitative comparisons for 2-view video generations. Each generation consists of two rows, where each row represents a sequence of generated frames, with columns showing frames at the same timestep. Neither the image nor the camera trajectories were used during model training. \textcolor{red}{Red} dotted lines are annotated to highlight object motion. Video results are included in the supplementary material for clearer comparisons.}
    \label{fig:mvvideo}
    \vspace{-4mm}
\end{figure}

\begin{table}[t]
    \centering
    \caption{Quantitative comparison for monocular geometry consistency on RealEstate10K test set.}
    \vspace{-2mm}
    \resizebox{0.8\textwidth}{!}{
    \begin{tabular}{c|cc|c|c|c}
     \toprule
        \multirow{2}{*}{Methods} & \multirow{2}{*}{FID $\downarrow$} & \multirow{2}{*}{FVD $\downarrow$} & \multirow{2}{*}{COLMAP error$\downarrow$} & Rot. AUC $\uparrow$ & Trans. AUC $\uparrow$  \\
        & & & & (@$5^\circ/10^\circ/20^\circ$) & (@$5^\circ/10^\circ/20^\circ$) \\
        \hline
        SVD & 16.89 & 139.64 & 30.3\% & 14.4 / 22.8 / 35.3  & 0.2 / 1.0 / 3.2  \\
        MotionCtrl & 21.09 & 119.06 & 55.0\%  &   8.6 / 13.9 / 22.2 & 0.6 / 2.1 / 5.7  \\
        CameraCtrl & 14.69 & 105.41 & 19.3\% &  21.4 / 32.9 / 48.4& 0.3 / 1.3 / 4.4 \\
        Ours & \textbf{11.43} & \textbf{55.10} & \textbf{14.4\%}  & \textbf{22.9 }/ \textbf{34.5} /\textbf{ 50.1} & \textbf{5.1} /\textbf{ 12.7} /\textbf{ 24.6} \\
    \bottomrule
    \end{tabular}
    }
    \vspace{-2mm}
    \label{tab:comparison}
\end{table}

\begin{table}[t]
    \centering
    \caption{Quantitative comparison for 2-view video generation.}
    \vspace{-2mm}
    \resizebox{0.85\columnwidth}{!}{
\begin{tabular}{c|c|cc|c|c|cc}
\hline
\multirow{2}{*}{Scenes} & \multirow{2}{*}{Methods} & \multirow{2}{*}{FID $\downarrow$} & \multirow{2}{*}{FVD $\downarrow$} & Rot. AUC $\uparrow$ & Trans. AUC $\uparrow$ & \multirow{2}{*}{Prec. $\uparrow$} & \multirow{2}{*}{MS. $\uparrow$} \\
 &   &  &  & (@$5^\circ/10^\circ/20^\circ$) & (@$5^\circ/10^\circ/20^\circ$) &  &  \\ \hline
\multirow{4}{*}{Real10K} &  SVD & 37.99 & 296.95 & 7.9 / 13.5 / 28.2 & 0.2 / 0.7 / 2.4 & 6.49 & 4.17 \\
 &  MotionCtrl & 29.23 & 277.05 & 8.1 / 16.5 / 29.4 & 1.5 / 5.3 / 16.1 &11.45  & 5.90 \\
 &  CameraCtrl & 12.57 & 131.32 & 22.4 / \textbf{{38.5}} / \textbf{{56.2}} & 0.6 / 2.5 / 8.2 & 19.49 & 11.25 \\
 &  Ours & \textbf{{8.82}} &\textbf{{94.86}} &\textbf{23.9} / 37.4 / 52.9 & \textbf{3.3} / \textbf{10.2} / \textbf{23.5} & \textbf{29.39} & \textbf{15.22}\\
\hline
\multirow{3}{*}{General} &  MotionCtrl  & 47.31 & 
 313.92 & 4.9 / 11.3 / 21.9 &  0.7 / 2.4 / 8.2 & 8.12 & 3.93\\
 &  CameraCtrl  & 26.71  & 221.23 & 14.1 / 26.9 / 43.2& 0.5 / 1.7 / 5.7 & 15.13 & 7.35 \\
 &  Ours  & \textbf{26.12} & \textbf{173.70} & \textbf{19.7} / \textbf{32.7 }/ \textbf{48.4} & \textbf{0.8 }/\textbf{ 2.8} /\textbf{ 8.7 }& \textbf{33.10} & \textbf{19.96}\\
\hline
\end{tabular}
    }
    \vspace{-4mm}
    \label{tab:multiview}
\end{table}

\paragraph{Multi-view Consistency} 
Alongside evaluating the individual monocular frame pose accuracy using COLMAP-based metrics, we further assess the cross-video consistency of the corresponding frames from generated multi-view videos.
We randomly sample 1,000 videos, each with 27 frames, from RealEstate10k~\citep{real10k} test set and convert each video into a two-view sequence with 14 frames per view. 
The new camera pose sequences are generated by setting the 14th frame as the world origin and positioning the remaining frames relative to it. The scales of the scenes are normalized so that the maximum distance from the origin is 1.
Following CVD~\citep{cvd}, we adopt SuperGlue~\citep{sarlin2020superglue} to find correspondences and estimate the camera poses between each time-aligned set of frames. 
SuperGlue not only measures angular errors in the rotation and translation but also computes the epipolar error of the matched correspondences. 
We similarly collect the AUC for frame pairs with rotation and translation errors below specific thresholds ($5^\circ$, $10^\circ$, $20^\circ$). The epipolar errors for the estimated correspondences are summarized to the precision (P) and matching score (MS). As shown in Tab.~\ref{tab:multiview}, our method outperforms baselines greatly. The ``Real10K'' category means that the input images are taken from the corresponding RealEstate10K test sequence, while the ``General'' means that the input images are taken from 1,000 randomly sampled images in the test split of our monocular video dataset.

 \looseness=-1
 \paragraph{Visual Quality} To assess the frame perceptual quality, we evaluate visual quality using FID~\citep{heusel2017gans} and FVD~\citep{unterthiner2018towards}. FID and FVD measure the feature-space similarity of two sets of images and videos, respectively. In our case, they quantify the distribution distance between the generated frame sequences and the ground-truth frames. We provide monocular evaluations in Tab.~\ref{tab:comparison} and multi-view evaluations in Tab.~\ref{tab:multiview}.
 As shown in these tables, our proposed framework enjoys the best visual quality. For both the ``Real10K'' and ``General'' categories, the ground-truth videos used to calculate these metrics are the video sequences corresponding to the input frames. These video sequences are from the test set split of the datasets and are not seen during training.

\subsection{Qualtitative Comparison}

We provide qualitative comparisons on RealEstate10k~\citep{real10k} scenes in Fig.~\ref{fig:mono} and text-to-image generated images in Fig.~\ref{fig:mvvideo}. As shown in Fig.~\ref{fig:mono}, our method produces videos with precise camera control, whereas MotionCtrl tends to generate overly smooth trajectories that simplify the viewpoint instructions, and CameraCtrl suffers from severe distortions at novel viewpoints. For example, in the first case, the camera instruction involves multiple panning operations, first panning left and then panning right. Still, MotionCtrl only pans left, ignoring the rest of the instructions. CameraCtrl's outputs, particularly in the first two cases, exhibit noticeable distortion, with the walls bending in the later frames.
Additionally, in the third and fourth cases, where the camera trajectories cover a long distance, both MotionCtrl and CameraCtrl produce unrealistic hallucinations, introducing artifacts such as merging indoor and outdoor pixels or distorting input pixels to compensate for a lack of generation ability.
 In Fig.~\ref{fig:mvvideo}, we observe that MotionCtrl and CameraCtrl tend to generate static scenes without any object motion. Although their methods produce realistic novel views, the synthesized objects remain static. In contrast, our method generates vivid object motion while maintaining accurate camera control. We highlight the object motion in  Fig.~\ref{fig:mvvideo} using auxiliary red lines. We encourage readers to view the supplementary videos for optimal visual comparisons.

\vspace{-2mm}
\subsection{Ablation Studies and Applications}
Due to the space limit, we refer readers to the Appendix for ablation studies and applications of our framework. We provide detailed ablation studies in Sec.~\ref{sec:ablation} on our proposed framework. Additionally, we explore the 3D reconstruction of our generated frames and four-view generation capabilities in Sec.~\ref{sec:application}. Videos are included in the supplementary material for optimal qualitative comparison.

\vspace{-2mm}
\section{Conclusion}

In this paper, we propose Cavia, a novel framework for consistent multi-view camera-controllable video generation. Our framework incorporates cross-frame and cross-view attentions for effective camera controllability and view consistency. Our model benefits from joint training using static 3D scenes and objects, animatable objects, and in-the-wild monocular videos. 
Extensive experiments demonstrate the superiority of our method over previous works in terms of geometric consistency and perceptual quality.

{\small 
\bibliographystyle{iclr2025_conference}
\bibliography{iclr2025_conference}
}

\clearpage
\appendix
\section{Additional Implementation Details}

Our training is divided into static stage and dynamic stage.
Our static stage is trained for around 500k iterations and our dynamic stage is trained for roughly 300k iterations.
The effective batch size is 128 and the learning rate is 1e-4. Our video length is 14 frames for each view with the first frame shared across views. Our model is fine-tuned at $256 \times 256$ spatial resolution from the SVD 1.0 checkpoint. The training data are prepared by first center-cropping the original videos and then resizing each frame to the shape of $256 \times 256$. In the dynamic stage, 30\% of iterations are used to train on monocular videos. During static training, the strides of frames are randomly sampled in the range of \texttt{[1, 8]}. For monocular videos, the strides are sampled in the range of \texttt{[1, 2]}. For dynamic multi-view object renderings, the strides are fixed to 1 to use all rendered frames since we already introduced randomness in the frame rate during rendering.
At inference time, the decoding chunk is set to 14 so all frames are decoded altogether. We sample 25 steps to obtain all our results.

\section{Additional Data Curation Details}
In this section, we provide additional details on our data processing and curation pipelines.

\paragraph{Static 3D Objects}
Our static objects data comprises multi-view images rendered from the Objaverse~\citep{deitke2023objaverse} and  Objaverse-XL\citep{deitke2023objaverse2} dataset. Similar to InstantMesh, we use a filtered high-quality subset of the original dataset to train our model. The filtering goal is to remove objects that satisfy any of the following criteria: (\romannum{1}) objects without texture maps, (\romannum{2}) objects with rendered images occupying less than 10\% of the view from any angle, (\romannum{3}) including multiple separate objects, (\romannum{4}) objects with no caption information provided by the Cap3D dataset, and (\romannum{5}) low-quality objects. The classification of “low-quality” objects is determined based on the presence of tags such as “lowpoly” and its variants (e.g., “low poly”) in the metadata. By applying our filtering criteria, we curated approximately 400k high-quality instances from the initial pool of 800k objects in the Objaverse dataset.

For each 3D object, we use Blender's EEVEE renderer to render an 84-frame RGBA orbit at $512\times 512$ resolution. we adaptively position the camera to a distance sufficient to ensure that the rendered object content makes good and consistent use of the image extents without being clipped in any view. For each frame, the azimuths can be irregularly spaced, and the elevation can vary per view. Specifically, the sequence of camera elevations for each orbit is obtained from a random weighted combination of sinusoids with different frequencies. The azimuth angles are sampled regularly, and then a small amount of noise is added to make them irregular. The elevation values are smoothed using a simple convolution kernel and then clamped to a maximum elevation of 89 degrees.

\paragraph{Static 3D Scenes}
Our static scenes data are sourced from RealEstate10k~\citep{real10k}, WildRGBD~\citep{xia2024rgbd}, MVImgNet~\citep{yu2023mvimgnet}, CO3Dv2~\citep{reizenstein2021common}, and DL3DV-10K~\citep{ling2023dl3dv}.
For RealEstate10k, we use the train/test split released by PixelSplat~\citep{charatan2023pixelsplat}. During training, we sample every 8 original frames to construct the training sequences.
For DL3DV-10K, we construct training sequences from the publicly available 7k subset. Since each video is very long for the DL3DV-10k dataset, we offline randomly sample multiple sequences from a single ground truth video to obtain multiple training data items.
For CO3Dv2, we remove the video sequences that contain whole-black images to avoid temporally inconsistent frames.
For WildRGBD and MVImgNet we use all classes available and removed sequences whose lengths are not enough for two-view training (shorter than 27 frames).

\paragraph{Dynamic 3D Objects} Our dynamic 3D objects are similarly rendered as the static 3D objects. The filtering pipelines remain mostly the same as the static objects, except that we introduce additional workflows to consider object motion. Inspired by previous works~\citep{liang2024diffusion4d,jiang2024animate3d,li2024vivid} that employ animatable objects from Objaverse. We render multiple fixed-view videos to examine the motion quality of the objects. We utilize lpips~\citep{zhang2018lpips} to measure the similarity of nearby frames and consider an object to be static if lpips similarity is above a certain threshold. Additionally, we render the alpha masks of the object and use this as an indicator of whether the object has moved out of the visible regions. Consequently, we remove objects with too large or sudden movements as well as objects with little-to-no motion. These filterings result in 19,000 objects. Our rendering strategy is also very similar to that of static 3D objects, introducing random elevation and azimuth changes to complicate the trajectories, except that we additionally introduce a random frame stride at rendering to augment the object motion. The stride is sampled individually for each object from the range \texttt{[1, 3]}. A larger the stride leads to renderings with faster object motion.

\paragraph{Monocular Videos} Our monocular video filtering pipeline involves filtering according to Particle-SfM output, OCR, aesthetic score, and camera motion. As mentioned in Sec.~\ref{sec:data}, we first attempt to annotate the camera poses for the video frames using Particle-SfM~\citep{zhao2022particlesfm}. Take InternVid~\citep{wang2023internvid} as an example, roughly 10 million video clips are processed and around 3 million samples are successfully processed by Particle-SfM. For each video, we start from the first frame and randomly select a frame stride of 1 or 2. The total number of images sent to Particle-SfM is 32 images. Our point count filtering is empirically implemented as a cut-off at 1,000 points and 40,000 points. Point clouds with too few points are removed due to the concern that the frames are poorly registered. Point clouds with too many points are avoided because their limited object motion. This aggressive filtering results in around 2 million samples for further processing. We then evaluate all the video clips using OCR detection algorithms and remove the samples whose detected text regions are larger then ${10^{-4}}$ of the image resolution (\textit{i.e.} 6 pixels). This process results in 604,000 samples. The next step is filtering with aesthetic scores and videos with aesthetic score annotations smaller than 4 are removed. 467,000 videos are left after these filtering process. Finally, we employ a camera motion classifier extended from the Open-Sora pipeline\footnote{\url{https://github.com/hpcaitech/Open-Sora/tree/main/tools/caption/camera_motion}}. The main motivation is that optical-flow on consecutive frames can be summarized to a global motion vector, assuming the most parts of the scene is moving in a uniform direction. Optical flow is first obtained using \texttt{cv2.calcOpticalFlowFarneback} for each consecutive frame pairs. Then, the magnitudes and directions are calculated via \texttt{cv2.cartToPolar}. These magnitudes and directions are classified into 8 categories: static, zoom out, zoom in, pan left, tilt up, pan right, tilt down, and unknown. The results of the frame pairs are summarized to obtain the final result of each video clip. When a certain type appears more than 50\%, the type for the whole video clip is determined directly. We aggressively classify a video clip as static if any of its frame pairs is categorized into static or unknown. Finally, we obtain 355,000 clips that satisfy our needs. The process is similarly applied to OpenVid~\citep{nan2024openvid}'s Panda-70M subset~\cite{chen2024panda} and we obtained 38,000 clips. In summary, our monocular video dataset consists of 393,000 clips.

\begin{table}[t]
    \centering
    \caption{Ablation Studies on each of our introduced modules. ``w/o Plucker'' refers to replacing the Plucker coordinate conditioning with one-dimensional conditioning as in MotionCtrl. ``w/o Cross-frame'' refers to replacing the Cross-frame attention with vanilla 1D temporal attention. ``w/o Cross-view'' refers to replacing the Cross-view attention with vanilla spatial attention. ``Ours (Static)'' means the model is only trained on static video datasets. ``Ours (w/o Mono)'' means that the model is fine-tuned on synthetic multi-view datasets, but is not trained with monocular video datasets. ``Ours (Full)'' means that the model is trained on all available data sources.}
    \resizebox{\columnwidth}{!}{
\begin{tabular}{c|c|cc|c|c|cc}
\hline
\multirow{2}{*}{Scenes} & \multirow{2}{*}{Methods} & \multirow{2}{*}{FID$\downarrow$} & \multirow{2}{*}{FVD$\downarrow$} & Rot. AUC $\uparrow$ & Trans. AUC $\uparrow$ & \multirow{2}{*}{Prec. $\uparrow$} & \multirow{2}{*}{MS. $\uparrow$} \\
&  &  &  & (@$5^\circ/10^\circ/20^\circ$) & (@$5^\circ/10^\circ/20^\circ$) &  &  \\ \hline
\multirow{4}{*}{Real10K} & w/o Plücker & 12.75 & 195.84 & 12.1 / 21.9 / 35.5 & 1.6 / 5.8 / 16.4 & 14.74 & 10.02 \\
 & w/o Cross-frame & 17.04 & 154.54 & 21.4 / 34.8 / 50.1 & 3.8 / 11.1 / 24.2 & 25.67 & 12.70 \\
 & w/o Cross-view & 9.45 &  106.82 &  22.8 / 36.7 / 52.4 &2.7 / 8.7 / 22.1 &27.57  & 14.65 \\
 & Ours & \textbf{8.82} & \textbf{94.86} & \textbf{23.9 }/ \textbf{37.4} / \textbf{52.9} & \textbf{3.3} / \textbf{10.2} / \textbf{23.5} & \textbf{29.39} & \textbf{15.22}\\
\hline
\multirow{5}{*}{General} & w/o Cross-frame & 71.39 & 249.02 & 9.8 / 19.1 / 32.7 & 0.5 / 1.9 / 6.6 & 13.20 & 8.97 \\
 & w/o Cross-view & 30.89 & 246.68 & 14.9 / 27.4 / 42.9  & 1.2 / 4.3 / 12.2 & 17.58 & 9.59  \\
 & Ours (Static) & 27.20 & 185.58 & 15.9 / 28.7 / 44.1 & 1.4 / 4.6 / 12.9 & 21.75 & 12.04 \\
 & Ours (w/o Mono)  & 35.79 & 243.05 &  15.0 / 27.1 / 42.6 & 0.3 / 1.3 / 4.2 & 18.55 & 10.78 \\
 & Ours (Full) & \textbf{26.12} & \textbf{173.70} & \textbf{19.7} / \textbf{32.7 }/ \textbf{48.4 }& \textbf{0.8 }/ \textbf{2.8 }/ \textbf{8.7} & \textbf{33.10} & \textbf{19.96} \\
\hline
\end{tabular}
    }
    \label{tab:ablation}
\end{table}

\begin{figure}[h]
    \centering
    \includegraphics[width=\linewidth]{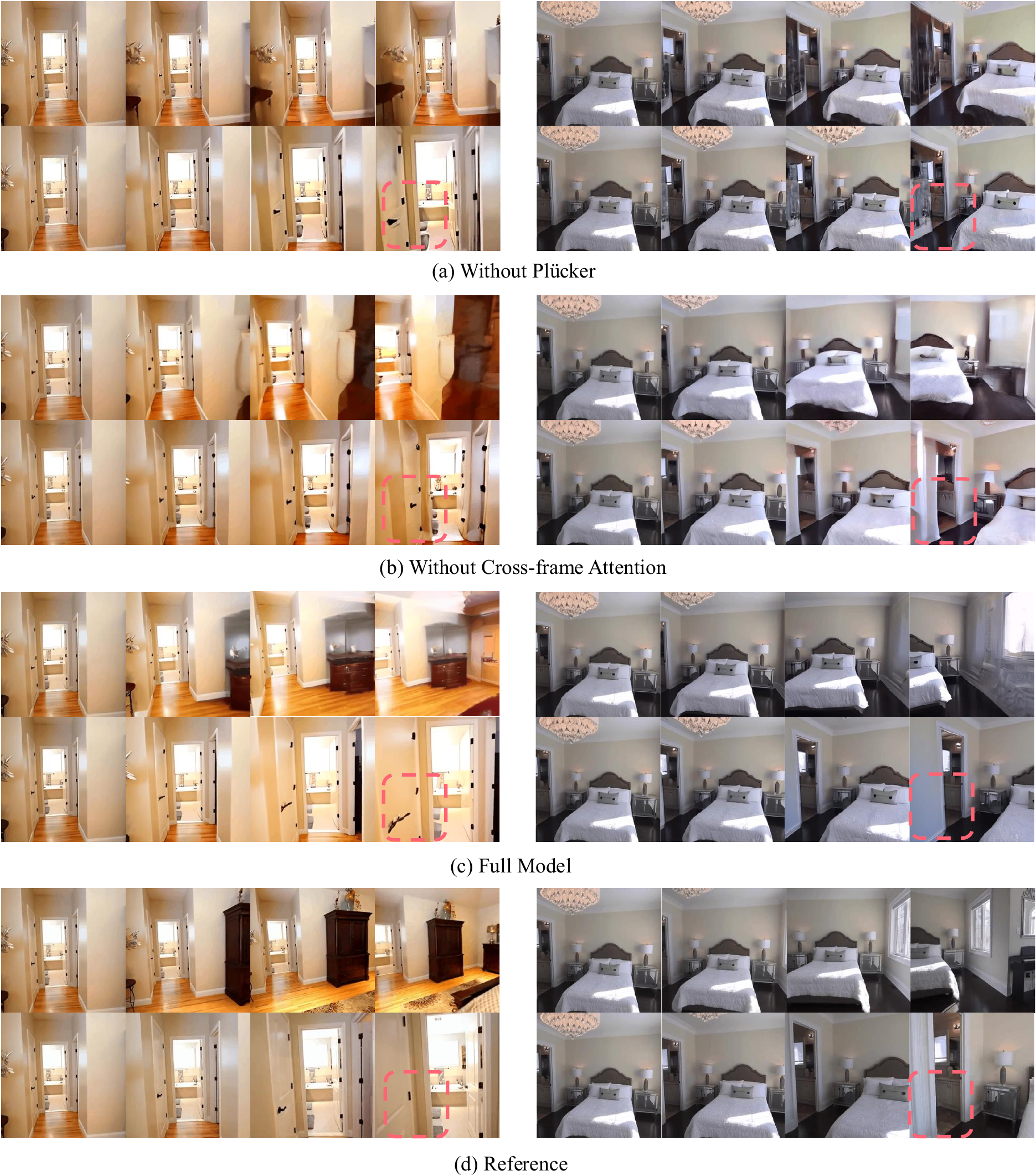}
    \caption{Ablation studies on Plücker coordinates and Cross-frame Attention. Video results are provided in supplementary for clearer qualitative comparisons.}
    \label{fig:ablation_static}
\end{figure}

\begin{figure}[h]
    \centering
    \includegraphics[width=\linewidth]{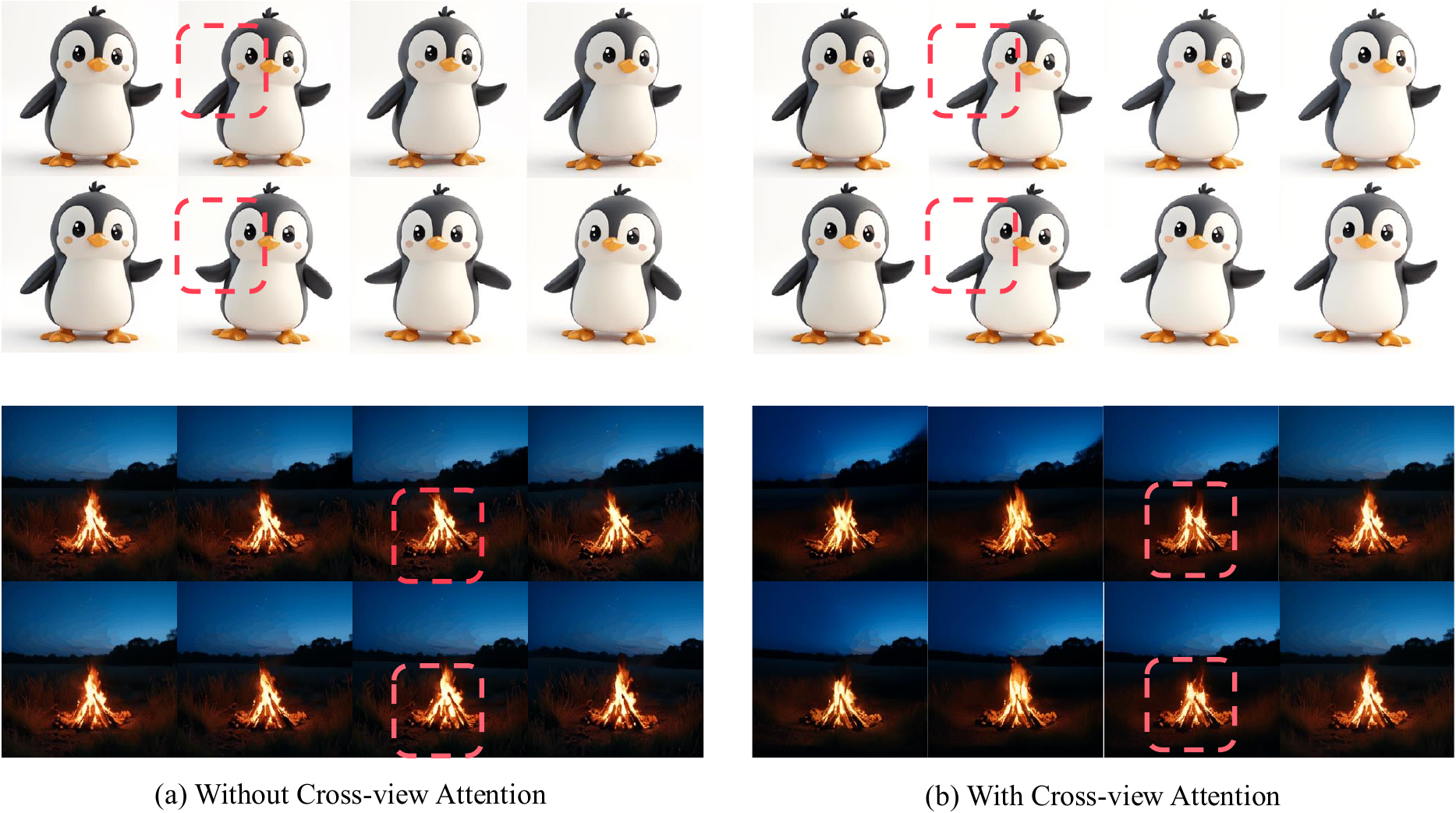}
    \caption{Ablation studies on Cross-view Attention. Video results are provided in supplementary for clearer qualitative comparisons.}
    \label{fig:ablation_crossview}
\end{figure}

\begin{figure}[h]
    \centering
    \includegraphics[width=\linewidth]{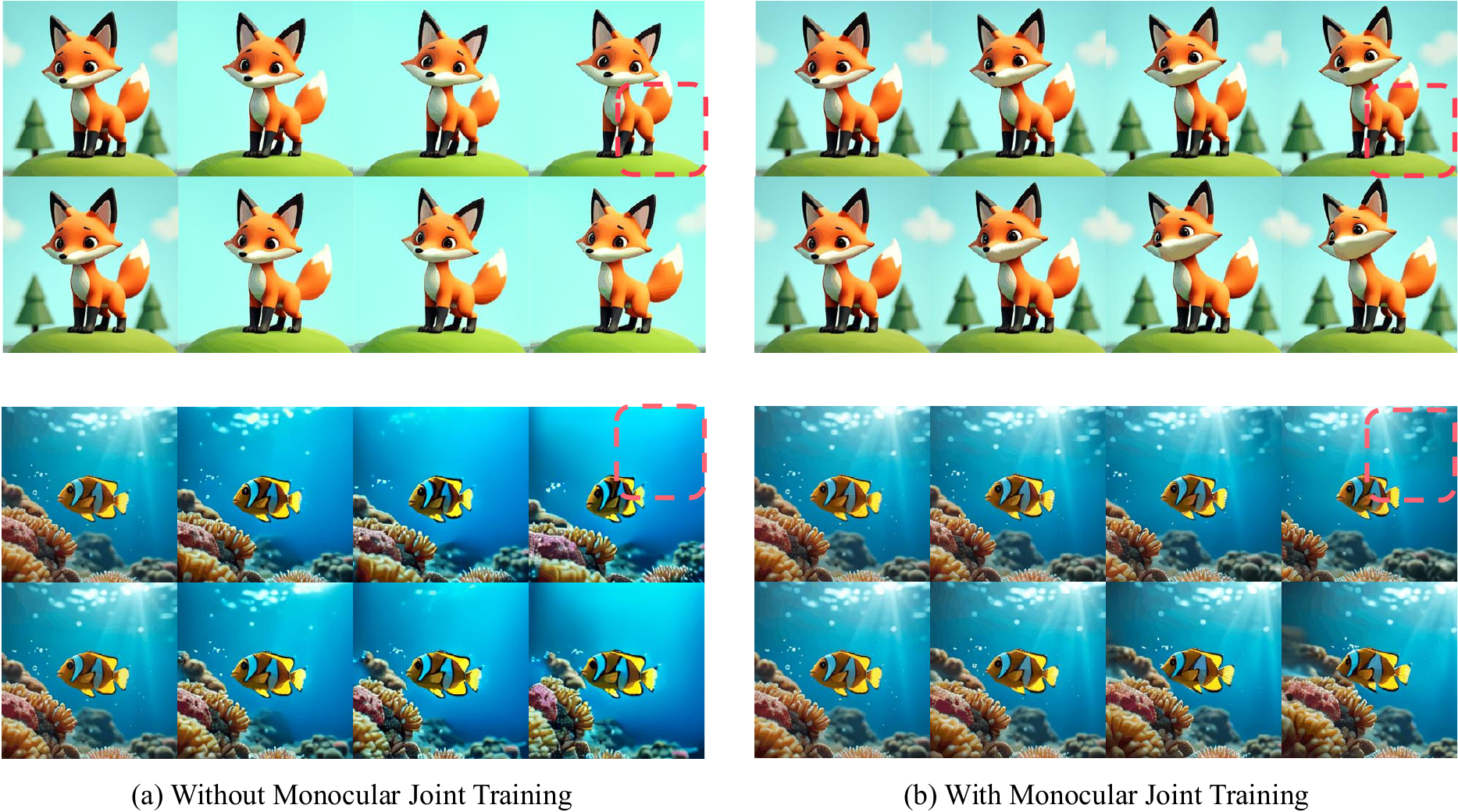}
    \caption{Ablation studies on the joint training strategy on monocular videos. Video results are provided in supplementary for clearer qualitative comparisons.}
    \label{fig:ablation_mono}
\end{figure}

\section{Evaluation Details}
For MotionCtrl and CameraCtrl, we use the open-source checkpoints trained from SVD released by the authors. These checkpoints are designed for image-to-video tasks so we can have fair comparisons. We use ``clean-fid''\footnote{\url{https://github.com/GaParmar/clean-fid}} and ``common-metrics-on-video-quality''\footnote{\url{https://github.com/JunyaoHu/common_metrics_on_video_quality}} for obtaining  FID and FVD, respectively. Our FVD results are reported in VideoGPT~\citep{yan2021videogpt} format. Our COLMAP is configured following DSNeRF~\citep{dsnerf} and CamCo~\citep{xu2024camco} to improve the few-view reconstruction performance. Concretely speaking, we enable \texttt{--SiftMatching.max\_num\_matches 65536} to support robust feature matching. To ensure that the SfM results best align with our videos, we set \texttt{--ImageReader.single\_camera 1} since most videos in our datasets consist of frames captured from a single camera.          

\section{Ablation Studies}
\label{sec:ablation}

In this section, we conduct extensive evaluations for ablation studies. We provide video comparisons in the supplementary.
We provide thorough quantitative comparisons in Tab.~\ref{tab:ablation} to illustrate the importance of our proposed components. The models are evaluated using RealEstate10K camera trajectories. For the ``Real10K'' and ``General'' categories, the testing images are from our test set split of RealEstate10K and InternVid, respectively.  Our full model enjoys the best perceptual quality and geometric consistency.

We first examine the importance of Plucker coordinates conditioning and the cross-frame attention modules. As shown in Fig.~\ref{fig:ablation_static}, model variants without cross-frame attention contains severe distortion artifacts, such as the bent walls. The model variant without Plucker coordinates results in simplified camera motion that ignores the complex camera viewpoint instructions.

We then evaluate the model variant without cross-view attention. As shown in Fig.~\ref{fig:ablation_crossview}, we observe that removing the cross-view attention module results in multiple individual video samples that contain different object motions. For example, the penguin moves differently in the first case, and the wood sticks in the fire appear differently in the second case. This behavior is not desirable because our goal is to obtain multiple videos from different camera paths of the same scene.

Finally, we examine the importance of our monocular video joint training strategy. As shown in Fig.~\ref{fig:ablation_mono}, we observe that when overfitting on dynamic objects from Objaverse, the generated results tend to contain frames with simplified backgrounds. This is mainly because, during the training, all data samples from Objaverse are implemented with single random color backgrounds. Our model benefits from joint training on monocular videos and preserves the ability to generate complex backgrounds when object motion is present.

\begin{figure}[t]
    \centering
    \includegraphics[width=\linewidth]{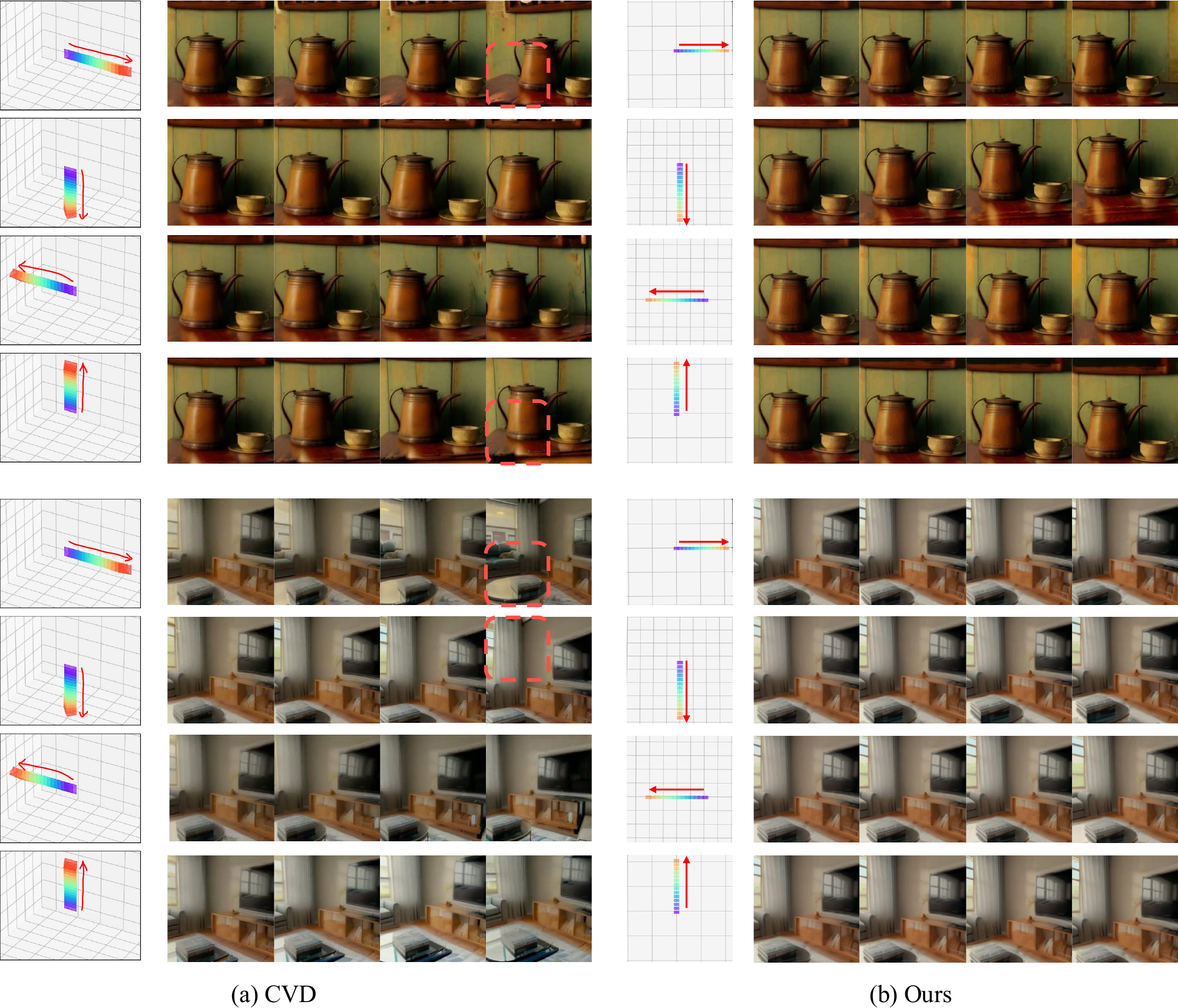}
    \caption{Four-view video comparison. The result of CVD is taken from their website. CVD tends to generate black border pixels, potentially due to its homography warping augmentations during training. In comparison, our method produces frames with better geometric consistency and perceptual quality.}
    \label{fig:more_views}
\end{figure}

\begin{figure}[h]
    \centering
    \includegraphics[width=\linewidth]{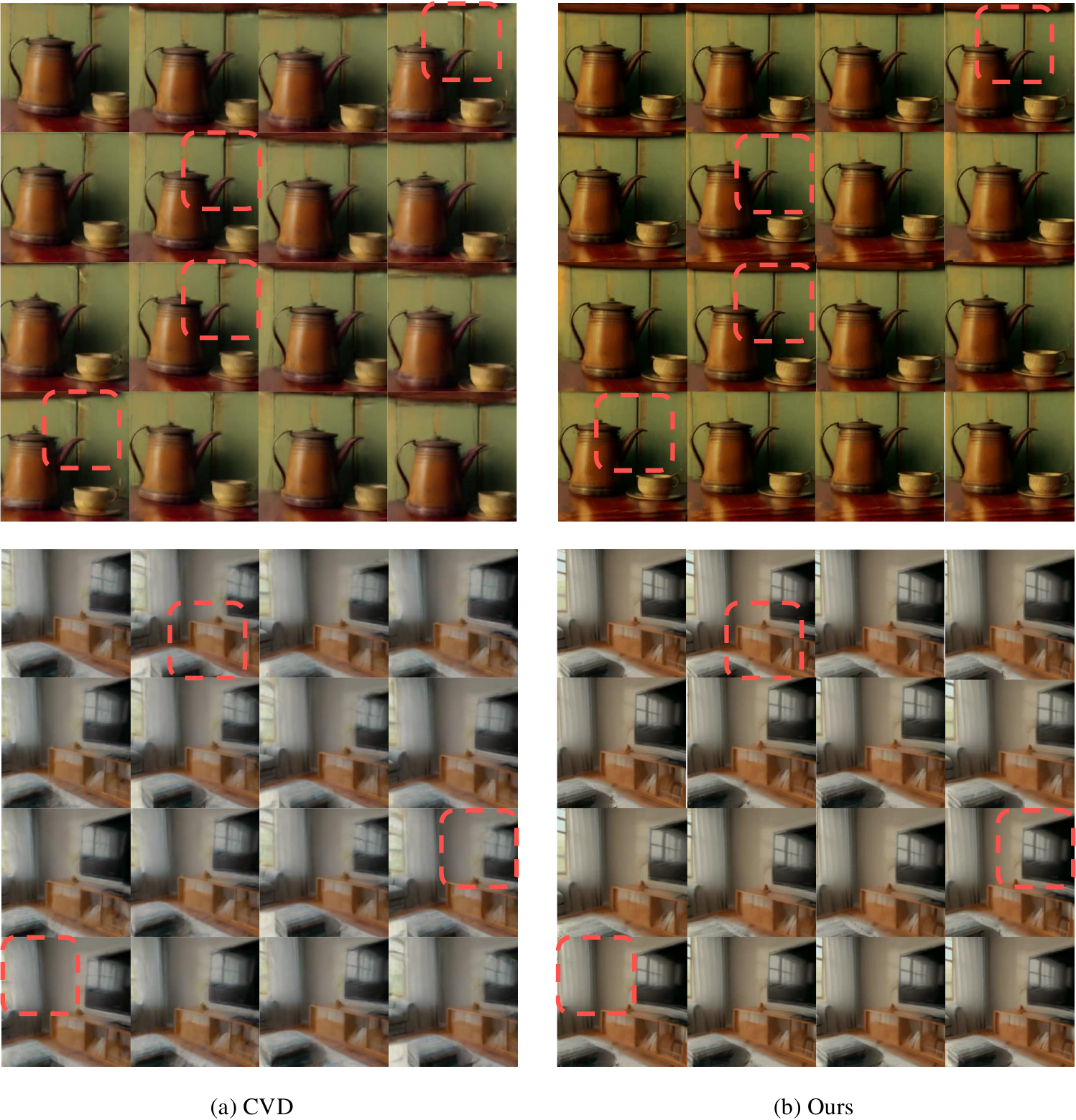}
    \caption{3D Reconstruction comparison. We render the reconstructed 3D Gaussians from an elliptical trajectory consisting of 16 novel views. The result of CVD is taken from their website. CVD's reconstruction results suffer from floaters and blurry artifacts due to the inconsistency in their generated frames. In comparison, our method produces sharper results with clearer visual quality.}
    \label{fig:3drecon}
\end{figure}

\section{Applications}
\label{sec:application}

In this section, we provide additional results on four-view inference and 3D reconstruction of our generated frames.
\subsection{Advancing to Four Views at Inference}

Our cross-view attention design enables us to extrapolate to more views straightforwardly at inference time. This design is more efficient compared with the concurrent work CVD~\citep{cvd} which requires enumeration of viewpoint pairs at inference time. We conduct a side-by-side comparison for 4-view generation in Fig.~\ref{fig:more_views}. Our method enjoys better consistency and shows more realistic results than CVD~\citep{cvd}. In comparison, CVD tends to produce artifacts at border regions. For example, the structure of the wall (first case) and the window (second case) change when the viewpoint changes. The results from CVD are taken from their author's website. We provide video comparisons in the supplementary. We also provide more 4-view generation results from Cavia in our supplementary.

\subsection{3D Reconstruction of Generated Frames}

We further perform 3D reconstruction on our generated frames. We render our reconstructed 3D Gaussians from an elliptical trajectory consisting of 16 novel views. We provide a side-by-side comparison with the concurrent work CVD~\citep{cvd} in Fig.~\ref{fig:3drecon}.
Compared with the results of CVD, our frames are more geometrically consistent and result in clearer 3D reconstruction and fewer floaters. For example, the results from CVD produce floaters on the cupboard regions and generate blurry artifacts for the wall and the TV due to inconsistencies. We provide video comparisons in the supplementary for clearer comparisons. We also provide additional 3D reconstruction results of Cavia's generated frames in the supplementary.

\section{Limitations}

Our framework has limited ability to generate large object motion, mainly due to the limitation of the base video generator SVD~\citep{svd}. We will explore better base models in future works.
Moreover, our data curation pipelines assume a simple camera model using shared camera intrinsic across frames. While enabling easier data preparation, this limits our model from generalizing to complex camera intrinsic changes at inference time, which is widely adopted in cinematography.
Additionally, for simplicity, our framework is trained with normalized scales of scenes, which can be further improved if potentially calibrated with metric scale. We will explore calibration techniques for better quality if a well-generalizable metric depth estimator becomes publicly available.

\end{document}

%% file: math_commands.tex
\usepackage{amsmath,amsfonts,bm}

\def\eqref#1{equation~\ref{#1}}

\def\1{\bm{1}}

\DeclareMathAlphabet{\mathsfit}{\encodingdefault}{\sfdefault}{m}{sl}
\SetMathAlphabet{\mathsfit}{bold}{\encodingdefault}{\sfdefault}{bx}{n}